\documentclass[letterpaper]{article}
\usepackage{aaai2026} 
\usepackage{times} 
\usepackage{helvet} 
\usepackage{courier}
\usepackage[hyphens]{url}
\usepackage{graphicx}
\usepackage{natbib} 
\usepackage{caption} 
\usepackage{algorithm}
\usepackage{algpseudocode} 
\usepackage{dsfont}
\usepackage{xcolor}
\usepackage{amsmath}

\newcommand{\blue}[1]{\textcolor{blue}{#1}}
\usepackage{newfloat}
\usepackage{listings}
\DeclareCaptionStyle{ruled}{labelfont=normalfont,labelsep=colon,strut=off} 
\floatstyle{ruled}
\newfloat{listing}{tb}{lst}{}
\floatname{listing}{Listing}
\usepackage{amsfonts}
\title{Early Lung Cancer Diagnosis from Virtual Follow-up LDCT Generation via Correlational Autoencoder and Latent Flow Matching}
\author{
    Yutong Wu \textsuperscript{\rm 1} \equalcontrib, 
    Yifan Wang \textsuperscript{\rm 1,2} \equalcontrib,
    Qining Zhang \textsuperscript{\rm 1}, 
    Chuan Zhou \textsuperscript{\rm 2},
    Lei Ying \textsuperscript{\rm 1}
}
\affiliations{
    
    \textsuperscript{\rm 1} Department of Electrical Engineering and Computer Science, University of Michigan, USA\\
    \textsuperscript{\rm 2} Department of Radiology, University of Michigan, USA\\
    \{ytongwu,wangyfan,qiningz\}@umich.edu, chuan@med.umich.edu, leiying@umich.edu
}

\usepackage{bibentry}
\usepackage{booktabs}
\usepackage{multirow}
\usepackage{pifont}

\begin{document}

\maketitle

\begin{abstract}
Lung cancer is one of the most commonly diagnosed cancers, and early diagnosis is critical because the survival rate declines sharply once the disease progresses to advanced stages. However, achieving an early diagnosis remains challenging, particularly in distinguishing subtle early signals of malignancy from those of benign conditions. In clinical practice, a patient with a high risk may need to undergo an initial baseline and several annual follow-up examinations (e.g., CT scans) before receiving a definitive diagnosis, which can result in missing the optimal treatment. Recently, Artificial Intelligence (AI) methods have been increasingly used for early diagnosis of lung cancer, but most existing algorithms focus on radiomic features extraction from single early-stage CT scans. Inspired by recent advances in diffusion models for image generation, this paper proposes a generative method, named \texttt{CorrFlowNet}, which creates a virtual, one-year follow-up CT scan after the initial baseline scan. This virtual follow-up would allow for an early detection of malignant/benign nodules, reducing the need to wait for clinical follow-ups. During training, our approach employs a correlational autoencoder to encode both early baseline and follow-up CT images into a latent space that captures the dynamics of nodule progression as well as the correlations between them, followed by a flow matching algorithm on the latent space with a neural ordinary differential equation. An auxiliary classifier is used to further enhance the diagnostic accuracy. Evaluations on a real clinical dataset show our method can significantly improve downstream lung nodule risk assessment compared with existing baseline models. Moreover, its diagnostic accuracy is comparable with real clinical CT follow-ups, highlighting its potential to improve cancer diagnosis.
\end{abstract}

\section{Introduction}

Lung cancer remains the most frequently diagnosed cancer worldwide, with nearly $2.5$ million new cases reported in 2022~\cite{bray2024global}. The prognosis of lung cancer is closely related to the clinical stage, and early diagnosis followed by timely treatments can significantly improve patients’ survival rate~\cite{goldstraw2016iaslc}. According to a multi-center cohort study reported in~\citet{he2022survival}, the 5-year overall survival rate drops sharply from $76.9\%$ at stage I to $21.4\%$ at stage IV, underscoring the crucial importance of diagnosing lung cancer at an early stage.

Currently, early diagnosis remains challenging because cancer symptoms are usually subtle at the early stages, and often appear only when the disease has progressed to an advanced stage. Low-dose CT (LDCT) has been recommended as a screening method to detect lung cancer in high-risk populations~\cite{cellina2023artificial}. Compared with standard CT, LDCT is less detrimental to patients’ health conditions with reduced radiation exposure. However, this benefit comes with high false positives and at the cost of multiple follow-ups for indeterminate pulmonary nodules detected at initial exam, making it more challenging to focus on timely, malignancy-related, subtle features at the early stage. In clinical practice, when early-stage LDCT does not provide sufficient evidence of malignancy, radiologists typically schedule follow-up examinations six to twelve months later so that a diagnostic decision can be made based on observed disease progression ~\cite{national2011reduced}. A patient with malignant nodules may have to wait for several months for follow-up LDCT scans to receive a definitive diagnosis, which delays diagnosis and treatment.

\begin{figure}
    \centering
    \includegraphics[width= 1\linewidth]{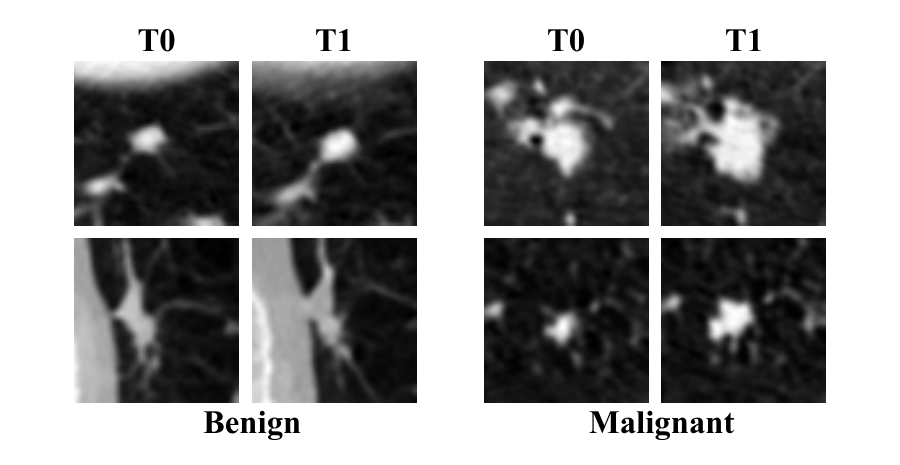}
    \caption{Examples of early (T0) and follow-up (T1) lung LDCT scans for benign and malignant lung nodules.}
    \label{sample image}
\end{figure}

As an illustrating example to illustrate how subtle the malignant signal could be, Figure~\ref{sample image} shows four early (T0) and follow-up (T1) lung LDCT scans from the \emph{National Lung Screening Trial} (NLST) dataset~\footnote{https://cdas.cancer.gov/nlst/}~\cite{national2011national}. Benign nodules typically remain stable, whereas malignant nodules usually exhibit noticeable growth over time. However, at an early stage, benign and malignant nodules are quite similar and difficult to distinguish.

Over the past decade, with the emergence of artificial intelligence (AI) approaches for the healthcare domain, numerous machine learning-based~\cite{gupta2024texture,liu2024lung} and deep learning-
based~\cite{yu2025etmo,ashames2025indifference,bbosa2024scs, wang2022hybrid} models have been developed for lung cancer diagnosis, aiming to classify lung nodules as malignant or benign accurately. However, most existing studies treated the diagnosis problem as a conventional single time point classification task, without considering the unique challenges of early diagnosis. This strategy overlooks the potential correlation between early and follow-up image features, as well as their dynamic changes that occur between the time of early prediction and the appearance of significant symptoms in follow-up exams, which inevitably leads to suboptimal prediction performance and potentially delayed diagnosis. To overcome this limitation, a study has modeled lung nodule progression as a Markov chain and applied reinforcement learning methods to facilitate earlier diagnosis by leveraging time-series information and formulating the task as a sequential decision-making problem \cite{wang2024leveraging, wang2024deep}.

Recently, diffusion models have demonstrated impressive performance in generating realistic images by learning to transform a prior (e.g. Gaussian distribution) into the target data distribution. These models could produce high-quality images with substantial diversity and have been adopted in medical imaging tasks, such as anomaly detection~\cite{kascenas2023role,fontanella2024diffusion}, medical image reconstruction~\cite{yang2023diffmic,mao2023disc,gao2023corediff}, image segmentation~\cite{chen2023berdiff,wu2024medsegdiff,rahman2023ambiguous}, and image synthesis~\cite{zhu2023make,dorjsembe2024conditional,meng2024multi}.

\begin{figure}
    \centering
    \includegraphics[width=0.9\linewidth]{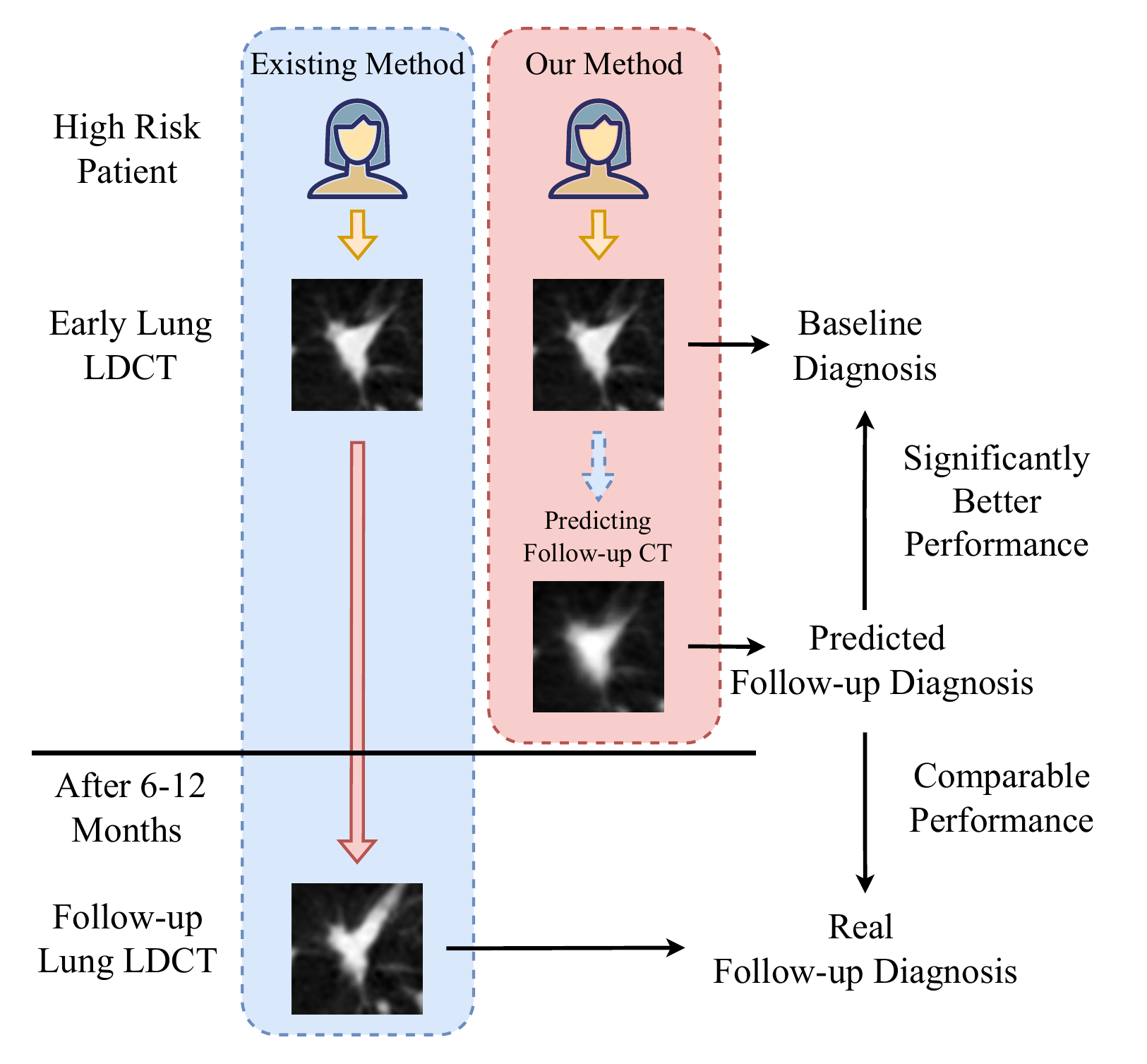}
    \caption{Comparison between existing lung cancer diagnosis methods and our proposed approach: unlike the clinical convention where a follow-up examination after 6–12 months is typically required to finalize the diagnosis, our method predicts the follow-up lung CT at an early stage, enabling more timely diagnosis for lung cancer patients.}
    \label{flow chart}
\end{figure}

Building on the strengths of diffusion models and aiming to address the need for early diagnosis while capturing the dynamic nature of lung cancer progression, we propose a diffusion-based framework (Figure~\ref{flow chart}) that generates follow-up lung nodule images from baseline LDCT scans to explore features that capture correlations and dynamic changes underlying nodule progression. The objective of our study is to utilize the generated follow-up nodule features to achieve performance comparable to real follow-up scans for improved diagnosis with baseline image alone, thereby facilitating early diagnosis without the need for long delays (e.g. one year in this paper).

\subsection{Our Contributions}

In this study, we address the challenge of early lung cancer diagnosis by predicting the temporal progression of lung nodules and generating a predicted one-year follow-up LDCT image. To capture the dynamic nature of disease progression, we formulate longitudinal nodule progression as a neural ordinary differential equation (ODE) in a latent space and frame the prediction of future follow-up LDCT scans as an image-to-image translation problem. We incorporate an auxiliary classifier during the training of the neural ODE to encourage the generated latent features to become more indicative of the original subtle malignancy-related characteristics. The latent space is learned by a correlational autoencoder, which serves as a shared temporal representation of the nodule, capturing disease progression-related patterns over time. It provides a more interpretable latent space that naturally distinguishes between progressive changes and static anatomy, while reducing image-specific noise from individual scans.

Our method, referred to as \texttt{CorrFlowNet}, consists of a correlational autoencoder and a latent space flow matching module, as described below:

\textbf{Correlational Autoencoder:} We design a correlational network~\cite{chandar2016correlational} based autoencoder, referred to as correlational autoencoder, to achieve representation learning and encode raw LDCT images into a latent space. It leverages not only the self-reconstruction loss to extract single-image features but also the cross-reconstruction loss along with feature space correlation loss to capture dependencies between early baseline and follow-up lung nodule features. These specifically designed loss functions enhance the correspondence between early baseline and follow-up LDCT embeddings, extracting features related to the dynamic progression of nodules manifested on the images from baseline to follow-up exams. To further improve the self-reconstruction performance and stabilize the training process, a normal autoencoder, commonly used in latent diffusion, is implemented as a pretrained warm start.

\textbf{Latent Flow Matching:} After constructing the latent space with the correlational autoencoder, we include a flow-matching model based on neural ODE, following the rectified flow framework, to map embeddings from baseline to those of follow-up LDCT scans. To improve training stability, we first incorporate a background alignment pre-training to learn invariant lung background structures that remain unchanged over time, allowing subsequent training to focus more effectively on the dynamics of the lung nodules. We then conduct flow matching using paired lung nodule images (baseline and corresponding follow-up LDCT). Finally, we apply an auxiliary classifier, leveraging pre-trained lung nodule classifiers to guide the neural ODE toward learning classification-oriented representations.

In summary, the main contributions of this paper include:
\begin{itemize}
    \item \textbf{Correlational Embedding:} We introduce a correlational autoencoder to better characterize the relationship between baseline and follow-up lung LDCT scans. We design a pre-training method to stabilize training for improving the self-reconstruction performance.
    \item \textbf{Classification Oriented Flow Matching:} We propose an auxiliary classifier-directed diffusion model training framework that penalizes the model when it exhibits low classification accuracy. The neural ODE–based diffusion model not only captures the relationship between early baseline and follow-up lung LDCT scans but also enhances its class-discriminative capability. 
    \item \textbf{Empirical Results:} We compare our method against several widely used diffusion and flow-matching methods. The results show that our approach not only outperforms existing methods in terms of early diagnosis accuracy, but also achieves diagnosis performance comparable to that obtained using real follow-up LDCT scans.
\end{itemize}

\subsection{Related Works}

The diffusion model has gained popularity recently due to its ability to produce high-quality images~\cite{dhariwal2021diffusion}. They have also been widely applied in medical imaging for disease diagnosis. Several methods directly applied diffusion for removing random noise and perturbation from medical images, leveraging diffusion models’ inherent denoising principle ~\cite{mao2023disc,gao2023corediff,jiang2025fast,xu2024stage}. 

For more advanced applications, some methods utilized diffusion to model the transformation between low-quality and high-quality medical images (e.g., low-dose CT and standard-dose CT). Other researchers further utilized the denoised image in some downstream tasks, such as anomaly detection~\cite{kascenas2023role} and image classification~\cite{yang2023diffmic}. With the development of conditional diffusion, more medical image tasks employed diffusion models, including image segmentation~\cite{chen2023berdiff,wu2024medsegdiff,rahman2023ambiguous}, image synthesis~\cite{zhu2023make,kim2024adaptive,meng2024multi}, and multi-modal fusion~\cite{yao2024addressing}. The key advantage of these methods is that conditional diffusion allows the integration of pathology or multi-modal data into the diffusion process, resulting in higher-quality, conditionally generated medical images.

Given the biologically stochastic and dynamically progressive nature of disease, and with the increasing availability of longitudinal medical imaging datasets, modeling disease progression has gradually attracted growing research attention. In~\cite{wang2024enhancing}, a U-Net was trained based on a Wasserstein Generative Adversarial Network framework to model the relationship between current and follow-up LDCT scans. In~\cite{liu2025imageflownet}, a model combining the U-Net structure and neural ODE/SDE framework is proposed. However, both models focused more on image quality, such as similarity with the ground truth and visual fidelity, rather than diagnosis performance. As in~\cite{liu2025imageflownet}, the image predictions were only tested based on image quality.

However, in the medical imaging domain, particularly for predicting future disease progression, high visual quality with artificial details does not necessarily translate into reliable diagnostic performance, as the biological system itself is inherently heterogeneous and uncertain. Our experiments reveal this trade-off between image fidelity and the preservation of class-relevant attributes.

\section{Methodology}
\begin{figure*}[ht]
    \centering
    \includegraphics[width=0.9\linewidth]{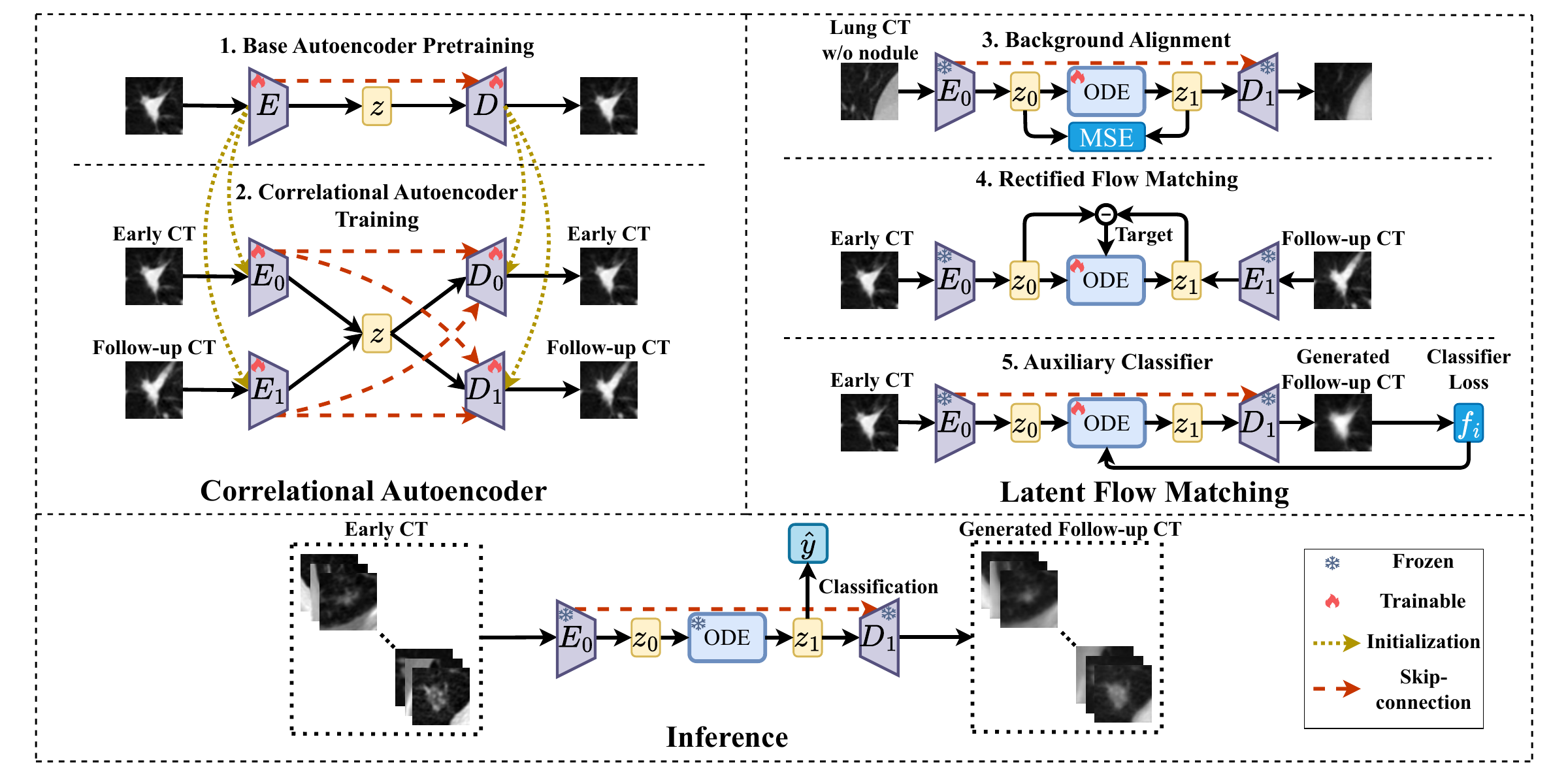}
    \caption{Overview of the proposed method. It consists of two modules: a correlational autoencoder and a latent flow matching module. In the correlational autoencoder, we first pre-train a base encoder-decoder pair ($E$, $D$) and then initialize the encoders $E_0$ and $E_1$ in the correlational autoencoder with the base encoder’s parameters $E$. Similarly, we initialize decoders in the correlational autoencoder $D_0$ and $D_1$ with $D$. In the latent flow matching stage, background alignment pretraining is performed, followed by rectified flow matching and the application of an auxiliary classifier loss $f_i$ to guide the neural ODE in learning class-specific attributes. During inference, the early baseline LDCT scan is used as input to generate follow-up-year CT embeddings $z_1$ for malignancy $\hat{y}$ prediction, along with a corresponding synthesized follow-up nodule image.}
    \label{pipeline}
\end{figure*}

In this paper, we aim to improve the early diagnosis of lung cancer by generating one-year virtual follow-up LDCT scans from baseline LDCT scans. The overview of our method is depicted in Figure~\ref{pipeline}, which consists of two modules: a correlational autoencoder and a latent flow matching module. At the first step, we develop a correlational autoencoder that is capable of capturing the relationship between early and follow-up LDCT scans, which encodes the LDCT image into a latent space. Second, we train a neural ODE in the latent space to model the longitudinal progression of the lung nodules with the help of an auxiliary classifier. Finally, during inference, we utilize the neural ODE to generate follow-up embeddings from early embeddings and predict nodule malignancy and reconstruct the corresponding follow-up nodule images to simulate a virtual follow-up, thereby enabling a potential earlier diagnosis. We present a simplified pseudo-code of our CorrFlowNet in Algorithm 1, with details provided subsequently.

\begin{algorithm}
\caption{CorrFlowNet}
\textbf{Input:} A training set including paired (early baseline and corresponding follow-up) lung LDCT images with nodule $(X_1, X_2)$ and without nodule $(H_1, H_2)$.
\label{alg:ModelFramework}
\begin{algorithmic}[1]
\State \textbf{Base Autoencoder Pre-train}
\Statex Initialize a base encoder-decoder pair $E$, $D$.
\Statex Train $E$ and $D$ by minimizing $\mathcal{L}_{\mathrm{BaseAE}}$ in Equation~\ref{equ_baseae} with $x\sim X_0 \cup X_1$.

\State \textbf{Correlational Autoencoder Training}
\Statex Initialize $D_0$ and $D_1$ with $D$, and $E_0$ and $E_1$ with $E$.
\Statex Train $D_0, D_1, E_0$, and $E_1$ by minimizing $\mathcal{L}_{\mathrm{CorrAE}}$ in Equation~\ref{equ_corae} with $(x_0, x_1) \sim (X_0, X_1)$.
\Statex Freeze $D_0, D_1, E_0$, and $E_1$.

\State \textbf{Background Alignment for ODE}
\Statex Initialize the neural ODE $v$ and generate embeddings $(h_0, h_1)$ from $(H_1, H_2)$ with $(E_0, E_1)$.
\Statex Train $v$ by minimizing $\mathcal{L}_{\mathrm{BA}}$ in Equation~\ref{Pre_ODE} with $(h_0, h_1)$.

\State \textbf{Rectified Flow Matching}
\Statex Generate embeddings $(z_0, z_1)$ from $(X_0, X_1)$ with $(E_0, E_1)$.
\Statex Train $v$ by minimizing $\mathcal{L}_{\mathrm{FM}}$ with $(z_0, z_1)$.

\State \textbf{Auxiliary Classifier Finetune}
\Statex Generate $\tilde{x}_1$ from $x_0 \sim X_0$ with Equation~\ref{x1t}
\Statex Finetune $v$ by minimizing $\mathcal{L}_{\mathrm{AC}}$ in Equation~\ref{AC}
\end{algorithmic}
\textbf{Inference:} $E_0, v, D_1$ forms a pipeline to generate the predicted follow-up nodule with early baseline as input. 
\end{algorithm}

\subsection{Notation}

We define the LDCT scan for patient $i$ at the early time point as $x_{0, i}\in R^{H\times W}$, and the corresponding follow-up scan as $x_{1, i}\in R^{H\times W}$, where $H\times W$ denotes the spatial dimensions of the LDCT scan. Given a total of $N$ patients, the collections of early and follow-up scans are represented as $X_0:= \{ x_{0, i} \}_{i=1}^{N}$ and $X_1:= \{ x_{1, i} \}_{i=1}^{N}$, respectively. Each scan has a ground truth label $y$ indicating whether the detected nodule is benign or malignant. We denote the set of all benign scans as $\mathcal{B}$ and all malignant scans as $\mathcal{M}$.

\subsection{Correlational Autoencoder}
To reduce the computational demands of training a flow matching diffusion model for image synthesis, 
we propose to use an autoencoder model that learns a latent space that can represent the image space, a design choice that has been widely adopted in other tasks ~\cite{rombach2022high}. However, as discussed previously, our early diagnosis task presents unique challenges that render a standard autoencoder insufficient. Due to the dynamic progression of lung cancer, relying solely on features that represent the early baseline LDCT image itself is inadequate for accurately predicting the follow-up progression and the true malignancy of a nodule. We require an encoder model capable of capturing features related to the dynamic changes of nodules between baseline and follow-up images, as these features reflect the progression of a nodule. Building on this idea, we develop a correlational autoencoder inspired by the principles of correlational neural networks~\cite{chandar2016correlational}. This model not only captures features representing the correlations between baseline and follow-up LDCT images but also preserves features intrinsic to the nodules themselves. As a trade-off, the correlational autoencoder is challenging to train. To stabilize training and accelerate convergence, we first pretrain a standard base autoencoder and then transfer its weights to initialize the correlational autoencoder.

\subsubsection{Base Autoencoder Pretraining} 
\label{subsub:basepre}
The pre-training process for the base autoencoder follows the standard procedure for representation learning, with only one encoder and one decoder. Given an image $x \sim X_0 \cup X_1$ in the pixel space, the encoder $E$ encodes $x$ into a latent space $z= E(x)$, where $z \in R^{h\times w\times c}$. The decoder reconstructs the latent embedding $z$ back into the pixel space as $\tilde{x} = D(z) = D(E(x))$, aiming to achieve high reconstruction quality that closely matches the original image $x$. The loss for the base autoencoder training (equation \ref{equ_baseae}) contains three components: pixel-wise $ L_2$ loss $\mathcal{L}_{L_2}$, deep perceptual loss $\mathcal{L}_{\mathrm{perc}}$, and a patch-based adversarial loss $\mathcal{L}_{\mathrm{adv}}$~\cite{isola2017image}. The combination of three losses encourages the model to preserve both pixel-level fidelity and perceptually meaningful high-level features during the reconstruction process:
\begin{equation}
\label{equ_baseae}
    \mathcal{L}_{\mathrm{BaseAE}} = \mathcal{L}_{L_2} + \mathcal{L}_{\mathrm{perc}} + \mathcal{L}_{\mathrm{adv}},
\end{equation}
where
\begin{equation}
\mathcal{L}_{L_2} = \mathbb{E}_{x \sim X_0 \cup X_1} \big[ \| x - D(E(x)) \|_2^2\big], \nonumber
\end{equation}
\begin{equation}
\mathcal{L}_{\mathrm{perc}} = \mathbb{E}_{x \sim X_0 \cup X_1} \Big[ \sum_{l} \| \phi_l(x) - \phi_l(D(E(x))) \|_2^2 \Big],\nonumber
\end{equation}
\begin{equation}
\mathcal{L}_{\mathrm{adv}} = \mathbb{E}_{x \sim X_0 \cup X_1} \big[ (DS(D(E(x))) - 1)^2 \big],\nonumber
\end{equation}
where $\phi_l(\cdot)$ represents the feature maps extracted from the $(l)$-th layer of a pretrained perceptual network~\cite{zhang2018unreasonable}, $DS$ denotes a discriminator trained in a manner similar to GANs~\cite{goodfellow2014generative}. It aims to distinguish generated images from real ones, providing a strong supervisory signal that helps prevent noisy or unrealistic image generation \cite{dosovitskiy2016generating}.

\subsubsection{Structure of Correlational Autoencoder}
The correlational autoencoder consists of two encoders $E_{0}$ and $E_{1}$, i.e., encoders for early and follow-up lung nodule CT, and two corresponding decoders $D_{0}$ and $D_{1}$, respectively. The idea behind this architecture is to encode different stages of the LDCT image into a correlated latent space. While $E_0$ and $E_1$ respectively take early and follow-up LDCT scans as input and encode them into the latent space, $D_0$ and $D_1$ take latent embeddings and respectively reconstruct the embeddings into early and follow-up LDCT scans. 

The two encoders share the same network architecture, a hierarchical convolutional neural network (CNN), and the decoder consists of a combination of upsampling layers and convolutional neural network structures. Since the latent space contains more information related to correlated features, it lacks sufficient low-level detail features as a trade-off. Inspired by the U-Net~\cite{ronneberger2015u}, we add skip connections at different levels from the encoder to the decoder. These connections transfer low-level details from the encoder layers to the decoder layers, helping to preserve image fidelity in the generated LDCT predictions. We validate the importance of these skip connections for improving image quality in our experiments.

The parameters of encoders and decoders of the correlational autoencoder are initialized from the pretrained base autoencoder. During training, the model is anticipated to perform both self-reconstruction and cross-reconstruction tasks (e.g., reconstructing follow-up LDCT images from baseline LDCT), which helps to capture features associated with the dynamic changes of nodules between baseline and follow-up images and reduce image-specific noise from individual scans.

\subsubsection{Correlational Autoencoder Training}

Using the pretrained encoder-decoder as a warm start, we initialize $E_{0}$ and $E_{1}$ with pre-trained weights of $E$, and $D_{0}$ and $D_{1}$ with pre-trained weights of $D$ to stabilize the training process and improve self-reconstruction quality. Next, we train the correlational autoencoder with a weighted sum of a self-reconstruction loss $\mathcal{L}_{\mathrm{sr}}$, a cross-reconstruction loss $\mathcal{L}_{\mathrm{cr}}$, and a correlation loss $\mathcal{L}_{\mathrm{corr}}$ to encourage more substantial alignment between the latent embeddings of early and follow-up CT scans, thereby improving the performance of the flow matching module. 

Given early lung CT scans $X_0$ and corresponding follow-up lung CT scans $X_1$, the loss function for the correlational autoencoder is defined as
\begin{equation}
\label{equ_corae}
    \mathcal{L}_{\mathrm{CorrAE}} = \lambda_1 \mathcal{L}_{\mathrm{sr}} + \lambda_2 \mathcal{L}_{\mathrm{cr}} + \lambda_3 \mathcal{L}_{\mathrm{corr}},
\end{equation}
with $\lambda_1, \lambda_2, \lambda_3$ are weights to balance the losses. As part of the $\mathcal{L}_{\mathrm{CorrAE}}$, the self-reconstruction loss is defined as
\begin{equation}
    \mathcal{L}_{\mathrm{sr}} = \mathcal{L}(X_{0},D_{0}(E_{0}(X_{0}))) + \mathcal{L}(X_{1},D_{1}(E_{1}(X_{1}))) ,\nonumber
\end{equation}
the cross-reconstruction loss is constructed similarly:
\begin{equation}
    \mathcal{L}_{\mathrm{cr}} :=  \mathcal{L}(X_{0},D_{0}(E_{1}(X_{1}))) + \mathcal{L}(X_{1},D_{1}(E_{0}(X_{0}))) ,\nonumber
\end{equation}
where $\mathcal{L} = \mathcal{L}_{\mathrm{BaseAE}}$, and the correlation loss can be formulated as
\begin{equation}
    \mathcal{L}_{\mathrm{corr}} := -\mathbb{E}_{(B_0,B_1) \sim (X_0,X_1)} \Big[corr(E_{0}(B_{0}),E_{1}(B_{1}))\Big],\nonumber
\end{equation}
where $corr$ represents the sample-based Pearson correlation coefficient within each training batch and $B_0$ and $B_1$ represent training batch sampled from $X_0$ and $X_1$ respectively.

\subsection{Latent Flow Matching}

With trained correlational autoencoder, we obtain an efficient and low-dimensional latent space, where the embeddings emphasize the dynamic changes of nodules between baseline and follow-up scans so that the nodule progression can be effectively characterized. In the latent space, we apply flow matching to translate early baseline LDCT embeddings into their corresponding follow-up LDCT embeddings with the help of an auxiliary classifier. We choose to use a neural ODE instead of an SDE, as SDEs incorporate random perturbations, which can hinder classifiers from providing accurate guidance to the velocity field. Moreover, inspired by the \emph{reflow} step in the rectified flow matching algorithm~\cite{liu2022flow}, and considering the nature of our dataset, which includes paired (baseline–follow-up) nodule LDCTs, we design our framework to generate follow-up CT predictions deterministically using rectified straight flows. We utilize a U-Net backbone to model the neural ODE, consisting of four downsampling convolution layers and four upsampling transposed convolution layers. The neural ODE takes a combination of LDCT embedding and time $t$ as input and outputs the velocity at that point in a velocity field~\cite{liu2022flow}. 

Given a pair of early CT scan $x_{0}\in X_0$ and follow-up CT scan $x_{1}\in X_1$ from the same patient, the rectified flow induced from $(z_{0}, z_{1}) :=(E_{0}(x_{0}),E_{1}(x_{1}))$ is an ordinary differentiable model (ODE) on time $t$:
\begin{equation}
    dz_{t} = v(z_{t},t)dt.\nonumber
\end{equation}
We conduct latent flow matching through the following steps: background alignment, rectified flow matching, and an auxiliary classifier fine-tuning step.

\subsubsection{Background Alignment}
In most cases, the background structures surrounding lung nodules, such as the lung pleura, remain unchanged across longitudinal CT scans, even though the nodules themselves may exhibit diverse changes over time. Therefore, we first pre-train the neural ODE on a large dataset of lung CT scans without nodules, enabling it to learn the general style and anatomical structures of the lung in LDCT images. This pre-training step enables the model to focus more effectively on the nodules themselves in subsequent training stages. Given the latent space embedding of an early lung CT scan $h_{0}$ and its follow-up lung CT scan $h_{1}$, denote its latent space embedding at time step $t$ to be 
\begin{align*}
    \tilde{h_{t}}:=h_{0}+\int_{0}^{t}v(\tilde{h_{s}},s)ds, \quad \tilde{h_0}:=h_0.
\end{align*}
We obtain $\tilde{h_{t}}$ through the Euler Solver and training of the neural ODE with the following loss:
\begin{equation}
\label{Pre_ODE}
   \mathcal{L}_{\mathrm{BA}} = \mathbb{E}\left[ \|h_1 - \tilde{h_1}\|_2^2 \right].
\end{equation}

\subsubsection{Rectified Flow Matching}
After background alignment, we perform rectified flow matching to learn the relationship between early and follow-up lung CT embeddings. We utilize rectified flow matching because it follows straight line paths as much as possible and solves both domain transfer between two distributions and generative modeling \cite{liu2022flow}. Given early and follow-up lung CT embeddings on the latent space $(z_{0}, z_{1})$. We train the neural ODE $v$ with the following loss:
\begin{equation}
\label{ODE}
\mathcal{L}_{\mathrm{FM}} = \mathbb{E}_{t \sim [0, 1]} \left[ \left\| z_{0} - z_{1} -  v\left(t z_{1} + (1 - t) z_{0}, t\right) \right\|^2 \right].
\end{equation}

\subsubsection{Auxiliary Classifier}

To encourage the diffusion model to capture more malignancy-relevant attributes, we add an auxiliary classifier, a combination of $m$ pre-trained \texttt{ResNet-50} lung nodule classification models ($f_i)~\cite{he2016deep} $, to finetune the neural ODE. 
Given the predicted follow-up CT scans 
\begin{equation}
\label{x1t}
    \tilde{x}_1:= D_1(\tilde{z}_1) \quad (\tilde{z_{t}}:=z_{0}+\int_{0}^{t}v(\tilde{z_{s}},s)ds, \quad \tilde{z_0}:=z_0),
\end{equation}
We obtain $\tilde{z_{t}}$ through the Euler Solver and train the neural ODE $v$ using the loss:

\begin{equation}
\label{AC}
\mathcal{L}_{\mathrm{AC}} = \mathbb{E}_{f\sim\{f_i\}_{i=1}^m}\left[\big(\mathds{1}_{\{x\in\mathcal{B}\}}-\mathds{1}_{\{x\in\mathcal{M}\}}\big)\,f(\tilde{x}_1)\right].
\end{equation}

\subsection{Inference}
After training the correlational autoencoder and neural ODE, we perform inference by generating predicted follow-up LDCT scans and providing early diagnoses based on the corresponding latent embeddings. Given an early baseline LDCT scan $x_0 \sim X_0$, we obtain its latent embedding $z_0 = E_0(x_0)$ and predict the follow-up embedding $\tilde{z_{1}}$ using Euler Solver \cite{liu2022flow} with 100 uniform steps to solve $(\tilde{z_{1}}:=z_{0}+\int_{0}^{1}v(\tilde{z_{s}},s)ds, \quad \tilde{z_0}:=z_0)$. A pretrained classifier is then used to predict malignancy based on $\tilde{z_{1}}$ and use $D_1(\tilde{z_{1}})$ to generate the predicted follow-up LDCT images at pixel space.

\section{Experiments and Results}
\subsection{Experiments}
The performance of our proposed \texttt{CorrFlowNet} was evaluated using a real clinical lung cancer screening dataset, the \emph{National Lung Screening Trial} (NLST)~\cite{national2011national}. The dataset, evaluation methods, and four widely recognized and representative generative models employed as model baselines for comparison are described below.

\subsubsection{Dataset}

The NSLT dataset is a multicenter trial involving 33 centers across the United States. The NLST dataset is the largest available lung cancer screening dataset and the most representative dataset to date in terms of the number of participated patients, diversity and potential for application across a wide range of populations.

With approved access permissions, we collected a data set from 1,226 subjects, each of whom had at least one lung nodule with a longest diameter between 4 and 30 mm on the baseline LDCT scan, clinically referred to as an indeterminate nodule, with more than one follow-up LDCT examinations. This selection of clinically indeterminate nodules allows for a more robust evaluation of early-diagnosis performance. For each screen-detected lung nodule, we utilized its annual LDCT scan pairs taken between the baseline (T0) and the first follow-up (T1), as well as between T1 and the second follow-up (T2) if available, as nodule image pairs centered at each nodule. A total of $1,121$ LDCT nodule image pairs were extracted from the $776$ subjects ($165$ with lung cancer and $611$ without) in the training/validation set. Both T0-T1 and T1-T2 image pairs were used for the training. Each image pair of a lung nodule, whether from T0 to T1 or from T1 to T2, was labeled according to NLST diagnosis results (malignant or benign). The remaining $450$ subjects with $450$ T0-T1 LDCT nodule image pairs ($397$ benign and $53$ malignant) were used as an independent testing set. The training, validation and test sets were split at the patient level to ensure that there was no patient overlap or data leakage between subsets.

\begin{table*}[htbp]
\renewcommand\arraystretch{1}
\begin{center}
\begin{tabular}{cccccccc}
\toprule
\textbf{Model} & \textbf{AUPRC} $\uparrow$ & \textbf{P-value} & \textbf{AUROC} $\uparrow$ & \textbf{P-value} &\textbf{LPIPS} $\downarrow$ &\textbf{SSIM} $\uparrow$ &\textbf{FID} $\downarrow$\\
\midrule
CorrFlowNet \blue{(Ours)}  & \textbf{0.351 ± 0.011} & / & \textbf{0.778 ± 0.004} & / & \textbf{0.301} & 0.477 & 84.15\\
Stable Diffusion  & 0.264 ± 0.031 & \textless 0.001 & 0.745 ± 0.021 & \textless 0.001 & 0.399 & 0.399 & 140.57\\
Rectified Flow  & 0.310 ± 0.038 & 0.007 &0.752 ± 0.016 & \textless 0.001 & 0.303 & 0.404 & \textbf{68.14}\\
BBDM  & 0.307 ± 0.025 & \textless 0.001 & 0.754 ± 0.016 & \textless 0.001 & 0.336 & 0.417 & 120.51\\
ImageFlowNet  & 0.325 ± 0.025 & 0.012 & 0.772 ± 0.019 & 0.352 & 0.421 & \textbf{0.486} & 211.86\\
\midrule
Early CT Ground Truth  & 0.305 ± 0.032 & 0.001 &0.751 ± 0.013 & \textless 0.001  & / & / & / \\
Follow-up CT Ground Truth  & 0.360 ± 0.032 & 0.383 &0.791 ± 0.015 & 0.033 & / & / & /\\
\bottomrule
\end{tabular}
\caption{Comparison of our model with real early and follow-up lung CT scans and baseline models. P-values are computed using our method as reference. LPIPS, SSIM, and FID are measured against the follow-up CT ground truth. $\uparrow$ represents a higher value signifies an improved performance and $\downarrow$ is the opposite; $\text{p-value} < 0.05$ denotes statistical significance.}
\label{result1}
\end{center}
\end{table*}

\subsubsection{Model Evaluation}
The baseline LDCT images were input to the trained model to generate follow-up nodule embeddings and images. The outputs were then fed into a pre-trained nodule classification module to predict nodule malignancy. The classification module consisted of $10$ \texttt{ResNet-50} models~\cite{he2016deep}, each was pre-trained on real baseline and follow-up LDCT scans. The mean and standard deviation across the 10 models were calculated as evaluation metrics to assess robustness.

Model performance was primarily evaluated based on diagnostic accuracy using the generated follow-up nodule embeddings and images. The \emph{Area Under the Precision-Recall Curve} (AUPRC) and the \emph{Area Under the Receiver Operating Characteristic Curve} (AUROC) were used for performance evaluation. As our dataset was randomly selected from NLST study, it is imbalanced in terms of the number of cancer and non-cancer cases, but reflects the real population prevalence of lung cancer. Since false negatives are more harmful than false positives in this context, we used AUPRC in addition to AUROC to highlight classification performance on positive cases. Statistical significance was assessed using the t-test method, with p-values less than 0.05 considered as statistically significant.

Besides the evaluation of diagnosis performance, we also quantified image quality and similarity between the generated follow-up nodule image and the real follow-up nodule image using \emph{Learned Perceptual Image Patch Similarity} (LPIPS), \emph{structural similarity index} (SSIM), and \emph{Fréchet inception distance} (FID).

\subsubsection{Baselines}
For comparison, the real baseline LDCT images and the corresponding real follow-up nodule images, referred to as the Early Lung LDCT Ground Truth and Follow-up Lung LDCT Ground Truth, respectively, were input to the same classification model, serving as a clinical baseline to demonstrate the clinical significance of our model. 

Furthermore, we adapted four widely recognized and representative generative models to generate follow-up LDCT images, compared with our CorrFlowNet approach. We retrained these methods using the same dataset, and the generated follow-up nodule images from these models were evaluated using the same classification model to ensure a fair comparison. Further implementation details of the model are provided in the Appendix.

\begin{itemize}
\item \texttt{Stable Diffusion}~\cite{rombach2022high} is a widely used and commercialized diffusion model that operates in the latent space, employing a modified VQ-VAE for perceptual image compression and a DDPM or DDIM as the latent diffusion model, enabling efficient and high-quality image generation. 
\item \texttt{Rectified Flow}~\cite{liu2022flow} is a neural ODE-based approach that learning the most straightforward and most efficient flow between distributions, providing a principled and continuous transformation. 
\item \texttt{BBDM}~\cite{li2023bbdm} employs a stochastic Brownian bridge process as its flow-matching framework to model distributional transitions. 
\item \texttt{ImageFlowNet}~\cite{liu2025imageflownet} employs a U-Net backbone but replaces skip connections with learnable velocity fields. It enhanced image detail by learning velocity-guided skip connections, allowing precise modeling of subtle structural changes. 
\end{itemize}

\subsection{Results and Discussion}

In this section, we first present the effectiveness of \texttt{CorrFlowNet} in improving early lung cancer diagnosis. We then conduct an ablation study to analyze the contribution of each component within  \texttt{CorrFlowNet}. Finally, we explore the trade-off between diagnostic performance and image quality, providing further insights into the benefits of the proposed correlational autoencoder.

\subsubsection{Early Diagnosis Performance}

Using the follow-up nodules predicted by our \texttt{CorrFlowNet} model as input, the nodule classification model achieves a test AUROC of $0.778 \pm 0.004$ and an AUPRC of $0.351 \pm 0.011$, which is significantly higher than using the real baseline nodule images as input (AUPRC of $0.305 \pm 0.032$, $\text{P-value} = 0.001$) and comparable to using the real follow-up nodules (AUPRC of $0.360 \pm 0.032$, $\text{P-value} = 0.383$). Note, the results shown in Table 1 are relatively lower than those reported by state-of-the-art models in the literature. This is mainly due to our use of relative simple \texttt{ResNet-50} models as the nodule classifier as the nodule classifier, without any task-specific adaptation for lung nodule diagnosis. The objective of this classifier is not to achieve peak performance, but rather to demonstrate the effectiveness of the CorrFlowNet model in generating ``virtual" follow-up nodules, without introducing bias from more advanced classification. Moreover, the datasets and patient inclusion criteria differ, as we excluded many ``easy cases". Nevertheless, our results demonstrate the clinical relevance of our model: by generating a ``virtual" appearance of nodules one year after the initial screening that is associated lung cancer risks, our model allows a better management of indeterminate nodules in clinical screening program, facilitating timely diagnosis of lung cancer without waiting for the next scheduled LDCT exam.

\begin{table*}[h]
\centering
\begin{tabular}{cccccc}
\toprule
\multirow{2}{*}{\shortstack{Base \\ Autoencoder}} & \multirow{2}{*}{ \shortstack{Correlational \\ Autoencoder}} & \multirow{2}{*}{\shortstack{Rectified Flow \\ Training}} & \multirow{2}{*}{\shortstack{Auxiliary Classifier \\ Training}} & \multirow{2}{*}{AUPRC} & \multirow{2}{*}{AUROC} \\
\\[0.5ex]
\midrule
\ding{51} & \ding{55} & \ding{51} & \ding{51}  & 0.325 ± 0.009 & 0.756 ±  0.005 \\
\ding{55} & \ding{51} & \ding{55} & \ding{55}  & 0.336 ± 0.006 & 0.776 ±  0.005  \\
\ding{55} & \ding{51} & \ding{51} & \ding{55} & 0.345 ± 0.008 & 0.777 ±  0.004 \\
\ding{55} & \ding{51} & \ding{51} & \ding{51}  & \textbf{0.351 ± 0.011} & \textbf{0.778 ±  0.004}\\
\bottomrule
\end{tabular}
\caption{Ablation study on the proposed modules.}
\label{result2}
\end{table*}

\begin{table}[ht]
\begin{center}
\begin{tabular}{cccccc}
\toprule
\textbf{$\boldsymbol{\lambda_2}$} & \textbf{PRC} $\uparrow$ & \textbf{ROC} $\uparrow$ & \textbf{LPIPS} $\downarrow$& \textbf{SSIM} $\uparrow$ & \textbf{FID} $\downarrow$\\
\midrule
0.2  & 0.302 & 0.771 & \textbf{0.305} & 0.463 & \textbf{78.14}\\
0.4  & 0.316 & 0.781 & 0.310 & \textbf{0.470} & 87.83\\
0.6  & 0.305 & 0.784 & 0.309 & 0.465 & 88.17\\
0.8  & \textbf{0.333} & \textbf{0.786} & 0.311 & 0.464 & 84.28\\
\bottomrule 
\end{tabular}
\caption{Performance with different cross-reconstruction loss weights ($\lambda_2$). PRC represents AUPRC, and ROC represents AUROC. LPIPS, SSIM, and FID are measured against the real follow-up CT image. $\uparrow$ represents a higher value signifies an improved performance and $\downarrow$ is the opposite. }
\label{trade off}
\end{center}
\end{table}

When compared with baseline models, as shown in Table~\ref{result1}, our method outperforms all baseline approaches in both AUROC and AUPRC. \texttt{Stable Diffusion} operates in the latent space using a \texttt{DDPM} as the latent diffusion approach. It generates images by starting from Gaussian noise conditioned on the early-stage nodule embedding, which makes it difficult to synthesize realistic temporal evolutions of nodules. Consequently, it performs poorly in our longitudinal nodule follow-up generation task, yielding the lowest diagnostic performance among all compared methods. \texttt{Rectified Flow} and \texttt{BBDM} both follow the flow-matching diffusion paradigm. The former aims to learn the straightest transformation path of ODE between the source and target distributions, whereas the latter introduces a Brownian Bridge (SDE) constraint to regularize the flow trajectory. The flow-matching approach encourages smooth and temporally consistent transitions from early baseline to follow-up LDCT embedding, which better aligns with our biological progression prediction task, leading to an improved performance compared to \texttt{Stable Diffusion}. However, since these methods utilize a conventional VAE as the autoencoder, their performance remains significantly lower than that of our proposed model with the correlational autoencoder. \texttt{ImageFlowNet} learns multiscale joint representation spaces and models stochastic flow fields using ODEs/SDEs in the hidden layers of U-Net backbone. This design enables it to capture progression-related dynamics at multiple levels, resulting in improved performance compared to \texttt{Rectified Flow} and \texttt{BBDM}. However, it requires learning several SDEs (four at different scales), which significantly increases computational cost and necessitates a relatively larger dataset to achieve optimal performance.

\subsubsection{Ablation Study}
To better understand the factors contributing to the improved performance, we conducted an ablation study by removing different modules from our model and replacing them with components from baseline comparison methods when applicable, as shown in Table 2. When replacing the correlational autoencoder with a base autoencoder (first vs. fourth row), the diagnosis performance degrades significantly, supporting our central hypothesis that the correlational autoencoder provides a more informative latent space for flow matching, thereby  improves early diagnosis. Moreover, incorporating the rectified flow training strategy and the auxiliary classifier each contributed uniquely to the improved diagnostic performance, the full model achieved the best overall results.

\subsubsection{Trade-off Between Classification Performance and Image Fidelity}

During the experiments, we observed that the method that achieved the highest classification accuracy did not necessarily produce the best image fidelity, highlighting a potential trade-off between diagnostic performance and visual quality. A subsequent experiment was conducted to further characterize this trade-off. During the training of the correlational autoencoder, as in equation~\ref{equ_corae}, we fixed $\lambda_1 = 1$ and adjusted the weighting factor $\lambda_2$ to balance the importance of correlation reconstruction and self-reconstruction.

Table~\ref{trade off} illustrates that as the cross-reconstruction loss weight ($\lambda_2$) increases, diagnostic performance improves whereas image fidelity decreases. Real LDCT scans inherently contain noise, such as artifacts from patient motion during screening, which can adversely affect the accuracy of nodule diagnosis. By assigning greater weight to the correlation component in the latent space encourages the model to focus on capturing features related to the dynamic progression between early and follow-up LDCT scans, rather than image-specific details from individual scans, which is also an advantage of our \texttt{CorrFlowNet} for medical imaging analysis.

\section{Conclusion}

In this paper, we proposed a framework for early lung cancer diagnosis that synthesizes a one-year follow-up nodule LDCT scan from an initial baseline scan, enabling early diagnostic assessment without the need for actual follow-up imaging. Our proposed model, \texttt{CorrFlowNet}, employs a correlational autoencoder to construct a latent space encoding dynamic nodule progression features, coupled with a flow-matching module augmented by an auxiliary classifier to synthesize follow-up nodule images. Our numerical results on the real-world NLST lung cancer clinical trial demonstrate that \texttt{CorrFlowNet} outperforms baseline generative models and, most importantly, achieves comparable performance to using real follow-up LDCT images. In future work, we aim to integrate additional clinically relevant data, including radiology reports from the early baseline year, to develop a multi-modality generative framework that further improves the model’s capability.

\noindent{\bf Acknowledgments:} The work of Qining Zhang and Lei Ying is supported in part by U.S.\ National Science Foundation (NSF) under grants 2134081, 2324769, 2331780; AFOSR grant FA9550-24-1-0002. The work of Yifan Wang and Chuan Zhou is supported in part by the National Institutes of Health (NIH) grant number U01CA216459. 

\bibliography{aaai2026}

\end{document}


\section{Appendix}

\subsection{Architecture and Training Details}
In this section, we will display details in architectures and training procedures of \texttt{CorrFlowNet}. We will demonstrate these details in the correlational autoencoder, latent flow matching and follow-up CT classification respectively. All the models are trained using a RTX4070-30GB Laptop.
\subsubsection{Correlational Autoencoder}
We present the structure of the encoder, decoder and discriminator in Table ~\ref{encoder}, Table ~\ref{decoder} and Table ~\ref{discriminator}. There's skip connections between the output of layer1, layer2 and layer3 of the encoder and their corresponding layers in the decoder.
\begin{table}[H]
\centering
\caption{Encoder structure}
\begin{tabular}{c|cccc}
\hline
\textbf{Layer} & \textbf{Channels} & \textbf{Kernel Size} & \textbf{Stride} & \textbf{Padding}\\
\hline
1 & 32  & 3 & 1 & 1 \\
2 & 64  & 4 & 2 & 1 \\
3 & 128 & 4 & 2 & 1 \\
4 & 4  & 4 & 2 & 1 \\
\hline
\end{tabular}
\label{encoder}
\end{table}

\begin{table}[H]
\centering
\caption{Decoder structure}
\begin{tabular}{c|cccc}
\hline
\textbf{Layer} & \textbf{Channels} & \textbf{Kernel Size} & \textbf{Stride} & \textbf{Padding}\\
\hline
1 & 128 & 3 & 1 & 1 \\
2 & 64 & 3 & 1 & 1 \\
3 & 32 & 3 & 1 & 1 \\
4 & 1 & 3 & 1 & 1 \\
\hline
\end{tabular}
\label{decoder}
\end{table}

\begin{table}[H]
\centering
\caption{Discriminator structure}
\begin{tabular}{c|cccc}
\hline
\textbf{Layer} & \textbf{Channels} & \textbf{Kernel Size} & \textbf{Stride} & \textbf{Padding} \\
\hline
1 & 64 & 4 & 2 & 1 \\
2 & 128 & 4 & 2 & 1 \\
3 & 256 & 4 & 2 & 1 \\
4 & 512 & 4 & 1 & 1 \\
5 & 1 & 4 & 1 & 1 \\
\hline
\end{tabular}
\label{discriminator}
\end{table}

We list the training details of base autoencoder and correlational autoencoder in Table ~\ref{base} and Table ~\ref{correlation}.

\begin{table}[H]
\centering
\caption{Training Details in Base Autoencoder}
\begin{tabular}{|c|c|}
\hline
\textbf{Parameter} & \textbf{Value} \\
\hline
Batch size  & 64 \\
Epoch & 300\\
Learning rate & 2e-4  \\
Optimizer & Adam  \\
\hline
\end{tabular}
\label{base}
\end{table}

\begin{table}[H]
\centering
\caption{Training Details in Correlational Autoencoder}
\begin{tabular}{|c|c|}
\hline
\textbf{Parameter} & \textbf{Value} \\
\hline
Batch size  & 64 \\
Epoch & 50\\
Learning rate & 2e-4  \\
Optimizer & Adam  \\
$\lambda_1$ & 1 \\
$\lambda_2$ & 1 \\
$\lambda_3$ & 0.1 \\
\hline
\end{tabular}
\label{correlation}
\end{table}

\subsubsection{Latent Flow Matching}
We present the structure of the neural ODE in Table ~\ref{ode}. There's skip connections between the first 3 convolution layers and their corresponding transpose convolution layers.
\begin{table}[H]
\centering
\caption{Neural ODE structure}
\begin{tabular}{ccccc}
\hline
\textbf{Layer} & \textbf{Channels} & \textbf{Kernel Size} & \textbf{Stride} & \textbf{Padding} \\
\hline
Conv1 & 32 & 3 & 1 & 1 \\
Conv2 & 64 & 3 & 2 & 1 \\
Conv3 & 128 & 3 & 2 & 1 \\
Conv4 & 256 & 3 & 2 & 1 \\
\hline
TConv4 & 128 & 3 & 2 & 1 \\
TConv3 & 64 & 3 & 2 & 1 \\
TConv2 & 32 & 3 & 2 & 1 \\
TConv1 & 4 & 3 & 1 & 1 \\
\hline
\end{tabular}
\label{ode}
\end{table}

We display the training details of the latent flow matching part in Table ~\ref{flow matching}. The three epochs and learning rates respectively represent the training procedure of Background Alignment, Rectified Flow Matching and Auxiliary Classifier.

\begin{table}[H]
\centering
\caption{Training Details in Latent Flow Matching.}
\begin{tabular}{|c|c|}
\hline
\textbf{Parameter} & \textbf{Value} \\
\hline
Batch size  & 64 \\
Epoch 1 & 5\\
Epoch 2 & 10\\
Epoch 3 & 20\\
Learning rate 1 & 2e-4  \\
Learning rate 2 & 2e-5  \\
Learning rate 3 & 2e-5  \\
Optimizer & Adam  \\
\hline
\end{tabular}
\label{flow matching}
\end{table}

\subsubsection{Follow-up CT Classification} In this part, we utilize a shallow neural network with a hidden layer of 256 neurons to classify lung CT on feature space prediction with epoch 30, batch size 64 and learning rate 1e-3.

\subsection{Additional Results}

\begin{figure*}
    \centering
    \includegraphics[width=1\linewidth]{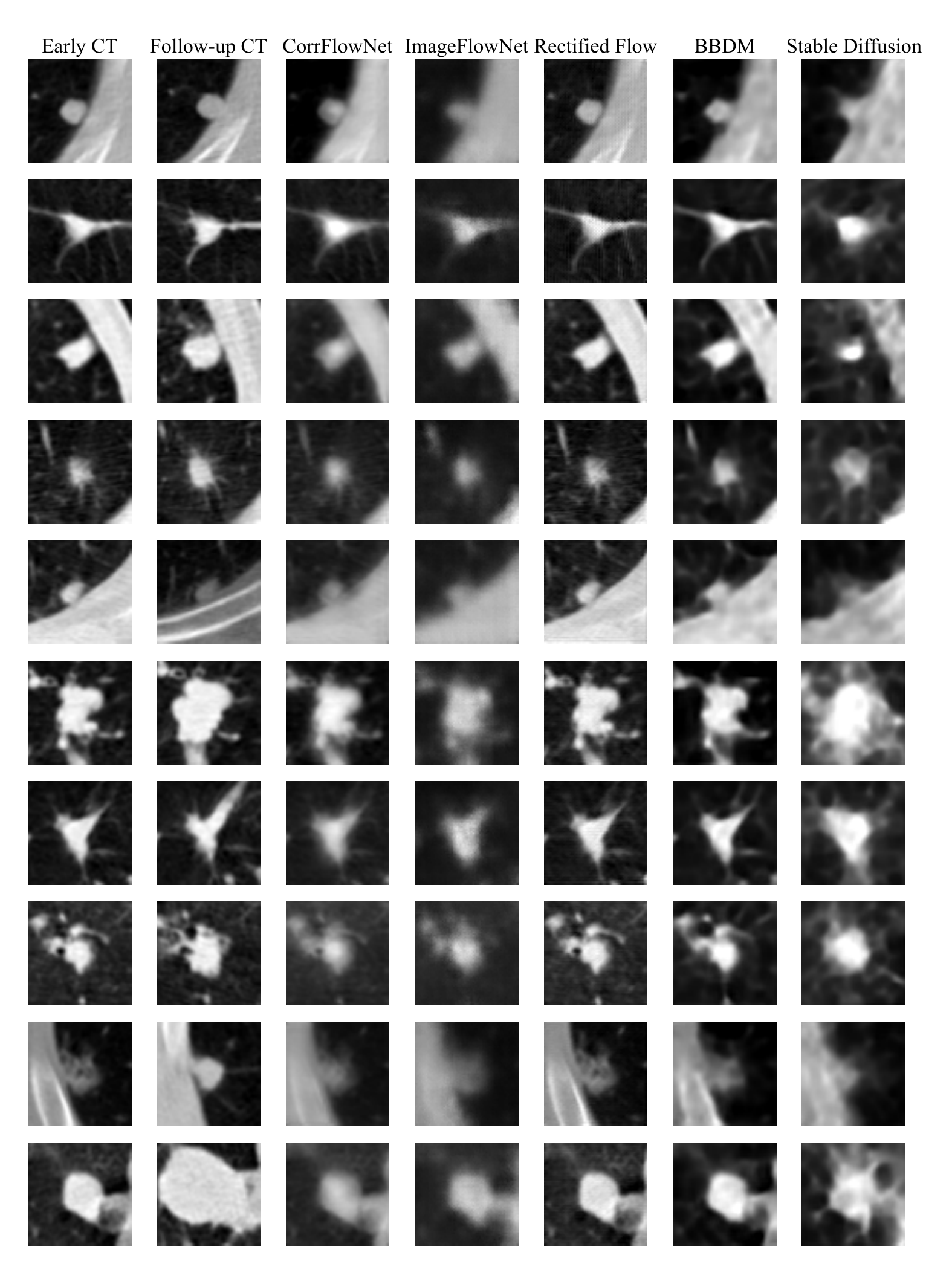}
    \caption{Examples of follow-up lung CT predictions generated by our \texttt{CorrFlowNet} with comparison to predictions from other baseline methods. The first five rows show two representative benign nodules, while the last five rows show two representative malignant nodules.}
    \label{image quality}
\end{figure*}

In this section, we will present follow-CT predictions generated with \texttt{CorrFlowNet}. We include early CT, follow-up CT ground truth, and follow-up CT predictions from other baseline methods for comparison in Figure ~\ref{image quality}.
